\documentclass[a4paper,twoside]{article}

\usepackage{epsfig}
\usepackage{subcaption}
\usepackage{calc}
\usepackage{amssymb}
\usepackage{amstext}
\usepackage{amsmath}
\usepackage{amsthm}
\usepackage{multicol}
\usepackage{pslatex}
\usepackage{algorithm2e}
\usepackage[bottom]{footmisc}
\usepackage{booktabs}
\usepackage{multirow}
\usepackage{graphicx}
\usepackage{SCITEPRESS}
\usepackage{tabularx}
\usepackage{xcolor}
\usepackage{subcaption}
\usepackage{hyperref}

\usepackage{multirow}
\usepackage{makecell} 

\begin{document}

\title{A Unified Framework for the Detection and Classification of Fatty Pancreas in Ultrasound Images}

\author{Ioan-Tudor-Alexandru Anghel\sup{1}, \authorname{Ciprian-Mihai Ceausescu\sup{1}, Elena Dana Nedelcu\sup{2}, Elena Raluca Știrban\sup{3}, Camelia Croitoru\sup{3}, Despina Ungureanu\sup{4}, Ana Maria Palan\sup{3}, Gabriela Pop\sup{5}}
\affiliation{\sup{1}Faculty of Mathematics and Computer Science, University of Bucharest, Bucharest, Romania}
\affiliation{\sup{2}Ponderas Academic Hospital, Radiology, Bucharest, Romania}
\affiliation{\sup{3}Ponderas Academic Hospital, Internal Medicine, Bucharest, Romania}
\affiliation{\sup{4}Ponderas Academic Hospital, Gastroenterology, Bucharest, Romania}
\affiliation{\sup{5}Emergency Clinical Hospital Bucharest,
Radiology Department, 
Bucharest}
\email{ioan-tudor.anghel@s.unibuc.ro, ciprian-mihai.ceausescu@drd.unibuc.ro, nedelcudana68@gmail.com}
}



\abstract{Non-alcoholic fatty pancreas disease (NAFPD) is an underdiagnosed condition associated with metabolic syndrome, insulin resistance, and increased risk of pancreatic cancer. Diagnosis typically relies on subjective visual assessment of ultrasound images by clinicians. We propose an end-to-end framework for automatically classifying normal versus fatty pancreas from abdominal ultrasound images. Our method employs a TransUNet-based segmentation architecture with a ResNet encoder and transformer bottleneck to delineate the pancreas and the splenic vein, followed by anatomically-guided patch extraction and patient-level classification through pairwise texture comparison. The feature engineering mimics clinical reasoning by comparing the echogenicity of peri-venous fat to the pancreatic parenchyma, providing an interpretable signal for classification. The segmentation models are initialized via domain-specific transfer learning from a liver segmentation task. We validate the full pipeline on a clinical dataset of 214 abdominal ultrasound images with 107 expert-labeled cases using 5-fold cross-validation. SVM with RBF kernel achieves a mean cross-validated accuracy of 89.7\%\,$\pm$\,1.8\% and F1 of 0.898\,$\pm$\,0.019, while the unsupervised K-Means baseline reaches 87.8\% accuracy, demonstrating that the proposed features capture the relevant clinical signal even without labeled training data. To our knowledge, this is the first end-to-end automated framework for fatty pancreas classification from ultrasound using segmentation-guided texture analysis.}

\onecolumn \maketitle \normalsize \setcounter{footnote}{0} \vfill

\section{\uppercase{Introduction}}
\label{sec:introduction}

Pancreatic steatosis, also known as non-alcoholic fatty pancreas disease (NAFPD), is a medical condition characterized by excessive fat infiltration in the pancreatic parenchyma, with significant implications for type 2 diabetes, obesity, cardiovascular diseases, and an increased risk of pancreatic adenocarcinoma \cite{smits2011pancreatic, tariq2019nafpd, Ahmad2026}. Although the clinical importance of NAFPD is recognized, its diagnosis remains challenging due to the subjective nature of conventional imaging evaluation, high inter-observer variability, and the anatomical position of the pancreas, which is difficult to visualize, relying on the subjective assessment of experienced radiologists \cite{lee2009fatty}.

The association between pancreatic lipomatosis and pancreatic neoplasms, including pancreatic ductal adenocarcinoma, one of the most lethal malignancies, with a 5-year survival rate below 10\%, suggests that pancreatic steatosis may represent not only a marker of metabolic dysfunction but also a potentially modifiable risk factor for pancreatic cancer prevention. The diagnosis of pancreatic steatosis in current clinical practice faces multiple methodological and technical challenges. Conventional evaluation using abdominal ultrasonography is inherently subjective and dependent on the radiologist’s experience, resulting in significant inter-observer variability.

Advanced imaging modalities, such as computed tomography (CT) and magnetic resonance imaging (MRI) \cite{Kuhn2015}, provide more objective methods for quantifying pancreatic fat content through CT attenuation measurements or proton density fat fraction (PDFF) \cite{Sakai2018}. However, these techniques require standardized acquisition protocols, are costly, involve exposure to ionizing radiation (in the case of CT), and are not universally available across all medical centers. Ultrasonography remains a practical, accessible, and cost-effective first-line imaging modality for detecting pancreatic lipomatosis, with severity grading providing clinically relevant prognostic information for metabolic risk stratification and early intervention \cite{Oh2021a, Oh2021b}. This supports its continued use in routine clinical screening and monitoring.

Most studies agree that pancreatic lipomatosis presents with increased echogenicity on ultrasonography and that severity classification (mild, moderate, severe) is clinically useful \cite{Ryu2023, Oh2021a, Oh2021b, Starodubova2019}. A new classification based on distinct echogenicity patterns has been proposed \cite{Keihanian2023}. Consistent ultrasonographic criteria, such as comparison with the renal cortex or retroperitoneal fat, are commonly used \cite{Starodubova2019, Oh2021a, Oh2021b}. In these studies, the authors apply different classification scales or reference limits; for example, some use hepatic echogenicity as a reference, while others use splenic echogenicity or local adipose tissue. The lack of a universally accepted classification system leads to variability in reported prevalence and severity.

Differences in ultrasonography equipment, operator experience, and echogenicity reference standards contribute to varied classifications. Deep learning models trained on ultrasound images may assist in the independent diagnosis of pancreatic steatosis \cite{Sun2024}. 

In this paper, we present a comprehensive framework that mimics the clinical reasoning process used by radiologists. Our main contributions are:
\begin{itemize}
    \item We propose the first end-to-end automated pipeline for fatty pancreas classification from ultrasound images, integrating segmentation, anatomically-guided patch extraction, and texture-based classification.
    \item We demonstrate effective transfer learning from liver to pancreas/splenic vein segmentation in ultrasound, and show that this domain-specific approach substantially outperforms zero-shot and fine-tuned MedSAM \cite{ma2024medsam} on this task.
    \item We introduce an anatomically-motivated feature engineering approach that compares peri-venous fat texture to pancreatic parenchyma texture, mimicking clinical assessment.
    \item We validate the full pipeline under 5-fold cross-validation with independent segmentation retraining per fold, demonstrating stable end-to-end classification performance (SVM RBF: 89.7\%\,$\pm$\,1.8\% accuracy, $\kappa{=}0.794\,\pm\,0.036$) and confirming that results are not an artefact of any particular data split.
\end{itemize}

\section{\uppercase{Related Work}}
\label{sec:related_work}

\subsection{Pancreas Segmentation in Medical Imaging}

\noindent Automated pancreas segmentation has been extensively studied in CT imaging. The U-Net architecture \cite{ronneberger2015unet} represents a foundational building block for most modern segmentation methods that were developed recently. In \cite{roth2018spatial}, the authors proposed a spatial aggregation approach using convolutional networks for automated pancreas localization and segmentation in CT volumes, on the three orthogonal axial, sagittal, and coronal views. To tackle the background ambiguity present in partially labeled datasets, \cite{zhou2019prior} proposes the integration of anatomical priors regarding abdominal organ sizes and fixed-point models for iterative refinement of organ segmentation. More recently, transformer-based architectures such as TransUNet \cite{chen2021transunet} and Swin-UNet \cite{cao2022swinunet} have achieved state-of-the-art performance by combining the local feature extraction capability of CNNs with the global context capturing of transformers. TransResU-Net \cite{tomar2022transresunet}, a variant architecture combining a ResNet encoder with transformer bottleneck layers and dilated convolutions, shows strong results on colonoscopy segmentation. However, pancreas segmentation in ultrasound remains largely unexplored due to the natural challenges of the investigation: speckle noise, low contrast boundaries, and high variability in pancreas appearance and position across patients.

\subsection{Fatty Pancreas Detection}

Fatty pancreas detection has been primarily studied using CT and MRI modalities. Quantitative approaches in CT typically measure Hounsfield unit attenuation values \cite{smits2011pancreatic}, while MRI-based methods use chemical-shift imaging to estimate fat fraction \cite{hu2010fat}. These approaches use objective measurements but they require expensive imaging that are not routinely used for pancreatic screening.

In ultrasound, fatty pancreas assessment remains largely qualitative. The clinical feasibility of ultrasound-based fatty pancreas grading but relied on manual visual assessment, has been demonstrated \cite{lee2009fatty}. Lately, a semi-quantitative scoring system based on echogenicity comparison has been developed \cite{sepe2011fatty}. To our knowledge, no fully automated pipeline exists for fatty pancreas classification from ultrasound.

\subsection{Transfer Learning in Medical Image Segmentation}

Transfer learning has proven effective for medical image segmentation, particularly when annotated data is scarce. Pre-training on natural images followed by fine-tuning on medical data is a common strategy \cite{tajbakhsh2016transfer, Patrascu2025}. Cross-organ transfer, where a model trained on one organ is fine-tuned for another, has also shown promise \cite{hosseinzadeh2021transfer}. Foundation models such as MedSAM \cite{ma2024medsam}, which adapts the Segment Anything Model (SAM) to medical imaging, have recently demonstrated strong generalization across diverse medical imaging tasks. The TransUNet family of architectures \cite{chen2021transunet, tomar2022transresunet} combines CNN encoders with transformer layers, enabling both local texture extraction and global context modeling. In our work, we use a variant \cite{tomar2022transresunet} that uses a ResNet50 \cite{He2016} encoder with a parallel transformer and dilated convolution bottleneck, which is suitable for capturing multi-scale information in ultrasound images.

\section{\uppercase{Methods}}
\label{sec:methods}

Our framework operates in three distinct stages, as follows:

\begin{enumerate}
    \item \textbf{Segmentation stage.} We employ a TransUNet-based architecture \cite{chen2021transunet, tomar2022transresunet} combining a ResNet \cite{He2016} encoder with transformer bottleneck layers to segment both the pancreas and the splenic vein from ultrasound images. The models are initialized via transfer learning from a liver segmentation task \cite{Xu2022} and fine-tuned on our clinical dataset.

    \item \textbf{Anatomically-Guided Patch Extraction stage.} Using the predicted segmentation masks, we extract tissue patches from two anatomically relevant regions: the pancreatic parenchyma (excluding the splenic vein) and the peri-venous fat region immediately beneath the splenic vein contour.

    \item \textbf{Classification via Texture Comparison stage.} For each patient, we compute per-patch texture features and derive a patient-level feature vector by summarizing pairwise distances between fat and pancreas patches. We evaluate both unsupervised (K-Means) and supervised (SVM, Random Forest, Logistic Regression) classifiers.
\end{enumerate}

\begin{figure*}[!ht]
  \centering
  \includegraphics[scale=0.159]{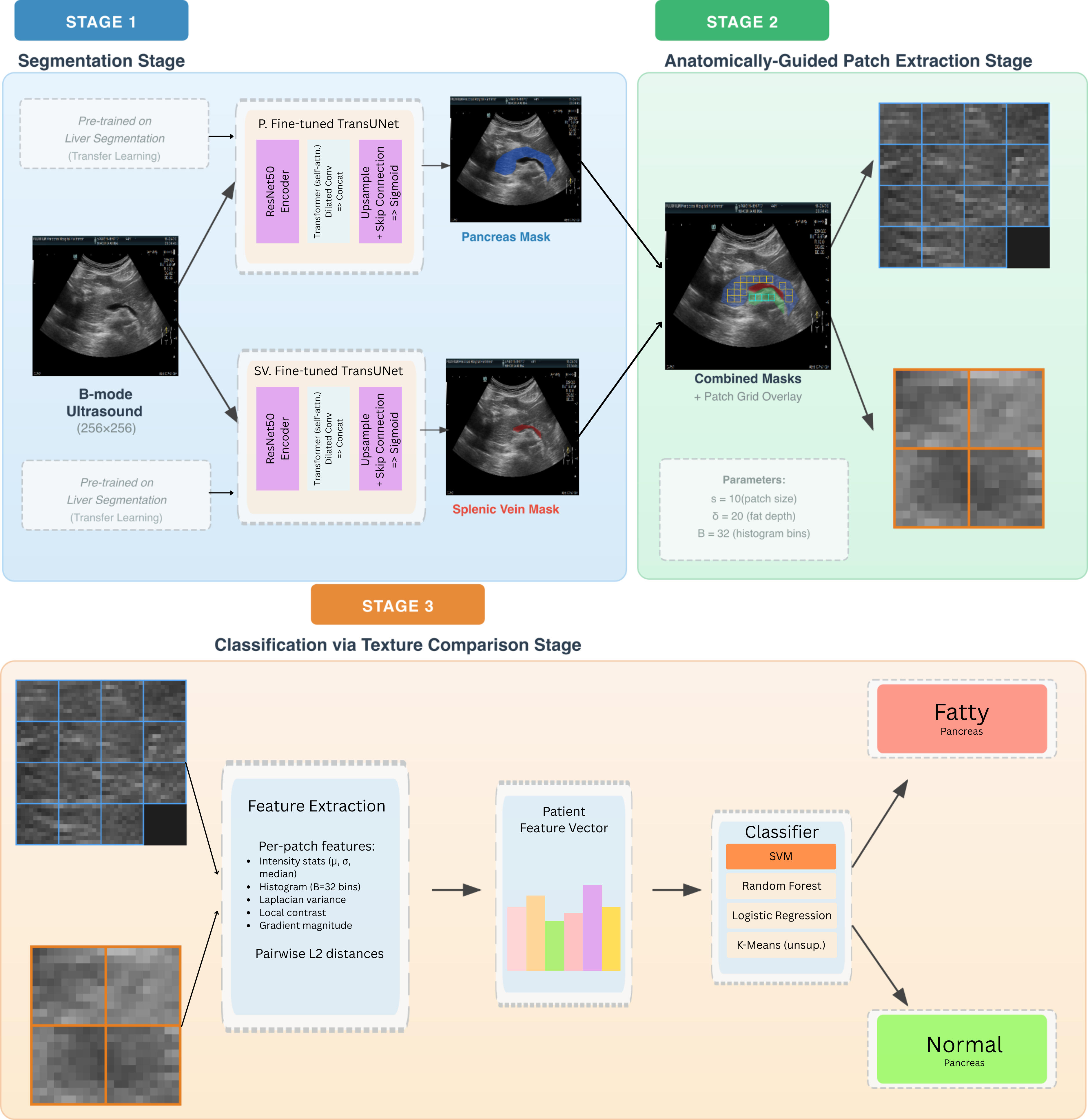}
  \caption{{\bf Overview of the proposed framework}. The pipeline takes a B-mode abdominal ultrasound image as input, segments the pancreas and splenic vein using TransUNet models (stage 1), extracts tissue patches from anatomically relevant regions (stage 2), computes pairwise texture features, and classifies the patient as having a normal or fatty pancreas (stage 3).}
  \label{fig:pipeline}
\end{figure*}

\noindent Figure~\ref{fig:pipeline} presents an overview of our proposed framework. Given an abdominal ultrasound image, the pipeline proceeds through three stages: segmentation of the pancreas and splenic vein, anatomically-guided patch extraction, and classification via texture feature comparison.

\subsection{Segmentation Architecture}

We employ a TransUNet-based architecture \cite{chen2021transunet} for semantic segmentation of the pancreas and splenic vein. The architecture consists of:

\textbf{Encoder.} A pre-trained ResNet50 backbone extracts hierarchical features at four resolution levels, where $H \times W$ denotes the input resolution:

\begin{equation*}
    \begin{aligned}
    \mathbf{F}_l &\in \left\{ \frac{H}{2} \times \frac{W}{2},\; \frac{H}{4} \times \frac{W}{4},\; \frac{H}{8} \times \frac{W}{8},\; \frac{H}{16} \times \frac{W}{16} \right\}, \\
    &where \quad l = 1, 2, 3, 4.
    \end{aligned}
\end{equation*}

\textbf{Bridge.} The bottleneck combines two parallel branches: (1) a Transformer encoder with multi-head self-attention that captures global dependencies across the feature map, and (2) a dilated convolution module with dilation rates $\{1, 3, 6, 9\}$ that captures multi-scale local context. The outputs are concatenated, yielding a feature representation that encodes both global and local information.

\textbf{Decoder.} Four decoder blocks progressively upsample the feature maps via bilinear interpolation and fuse them with skip connections from the corresponding encoder levels through residual blocks. A final $1 \times 1$ convolution with sigmoid activation produces the binary segmentation mask.

Two separate models are trained: one for pancreas segmentation and one for splenic vein segmentation. Both models are trained using a combined Dice and Binary Cross-Entropy (BCE) loss:

\begin{equation}
    \mathcal{L} = \mathcal{L}_{\text{BCE}} + \mathcal{L}_{\text{Dice}}
\end{equation}

\noindent where the Binary Cross-Entropy loss is defined as:

\begin{equation}
    \mathcal{L}_{\text{BCE}} = -\frac{1}{N} \sum_{i=1}^{N} \left[ g_i \log(p_i) + (1 - g_i) \log(1 - p_i) \right]
\end{equation}

\noindent and the Dice loss is defined as:

\begin{equation}
    \mathcal{L}_{\text{Dice}} = 1 - \frac{2 \sum_i p_i g_i + \epsilon}{\sum_i p_i + \sum_i g_i + \epsilon}
\end{equation}

\noindent with $p_i$ and $g_i$ denoting the predicted and ground-truth values for pixel $i$, respectively, and $\epsilon$ a smoothing constant for numerical stability.

\subsection{Transfer Learning Strategy}

Training deep segmentation models on small medical datasets is challenging due to the overfitting risk. We address this transfer learning as follows:

\begin{enumerate}
    \item \textbf{Pre-training on liver segmentation.} The TransUNet model is first trained on a liver segmentation dataset, which shares similar imaging characteristics (ultrasound, abdominal organs) but has more available annotated data.

    \item \textbf{Fine-tuning on target task.} The pre-trained weights are loaded, and the full model is fine-tuned with a reduced learning rate ($10^{-5}$ vs. $10^{-4}$) on the target dataset (i) Pancreas (labeled P. in results), and (ii) Splenic Vein (labeled S.V. in results). All layers are trainable during fine-tuning.
\end{enumerate}

Data augmentation is applied during training, including random rotation ($\pm 35^{\circ}$), horizontal and vertical flips, and coarse dropout, each with probability 0.3.

\subsection{Anatomically-Guided Patch Extraction}
\label{sec:patch_extraction}

\begin{figure*}[!h]
  \centering
  \begin{subfigure}[b]{0.75\columnwidth}
    \includegraphics[width=\linewidth]{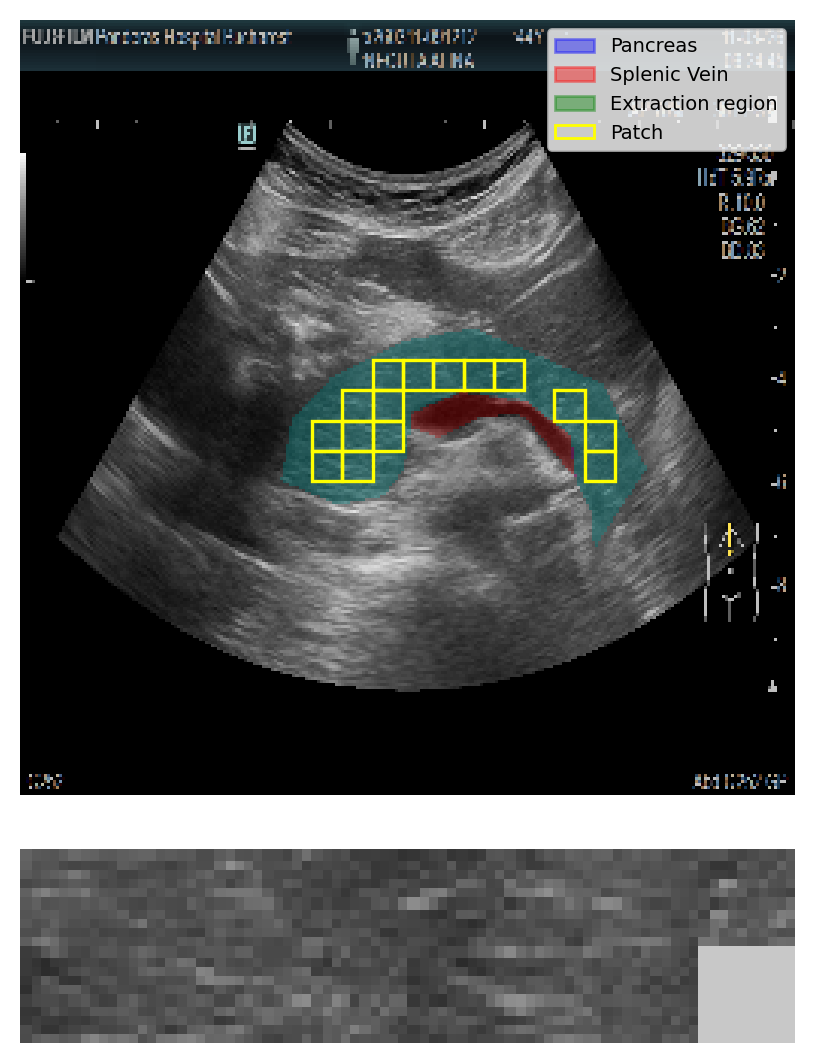}
    \caption{{\it Pancreas region: grid-based extraction within the pancreas mask (blue), excluding the splenic vein (red)}}
  \end{subfigure}
  \hspace{1cm}
  \begin{subfigure}[b]{0.75\columnwidth}
    \includegraphics[width=\linewidth]{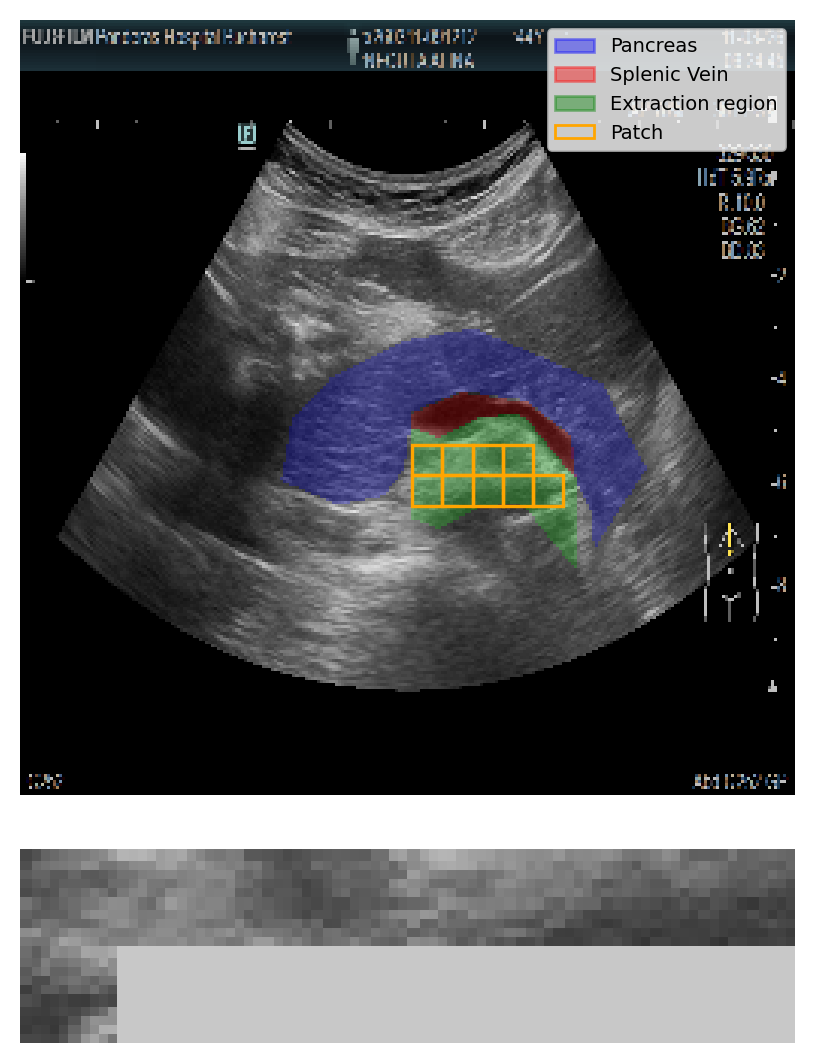}
    \caption{{\it Fat region: contour-following extraction below the splenic vein bottom contour, extending $\delta$ pixels downward.}}
  \end{subfigure}
  \vspace{0.3cm}
  \caption{{\bf Patch extraction strategy}. The top panels show the ultrasound image with segmentation masks, extraction regions (green), and patch locations (yellow/orange rectangles). Bottom panels show the extracted patches upscaled for visibility.}
  \label{fig:patch_extraction}
\end{figure*}

The clinical assessment of fatty pancreas involves comparing the echogenicity of the pancreatic tissue to the retroperitoneal fat surrounding the splenic vein. Our patch extraction strategy automates this comparison by sampling patches from two distinct regions, for a consistent quantitative evaluation, as illustrated in Figure~\ref{fig:patch_extraction}.

\textbf{Pancreas Patches.} We extract all non-overlapping patches of size $s \times s$ pixels from within the predicted pancreas mask, excluding any region overlapping with the splenic vein mask (Figure~\ref{fig:patch_extraction}, top). The grid-based extraction ensures complete coverage of the pancreatic parenchyma.

\textbf{Fat Patches.} We extract patches from the peri-venous fat region immediately beneath the splenic vein (Figure~\ref{fig:patch_extraction}, bottom). For each column $x$ in the splenic vein mask, we identify the bottom-most vein pixel $y_{\text{bottom}}(x)$ and define a valid extraction zone extending $\delta$ pixels below:

\begin{equation}
    \mathcal{R}_{\text{fat}} = \{(x, y) \mid y_{\text{bottom}}(x) < y \leq y_{\text{bottom}}(x) + \delta\}
\end{equation}

\noindent Patches are extracted in a non-overlapping grid from $\mathcal{R}_{\text{fat}}$, excluding any overlap with the pancreas or splenic vein masks. This contour-following strategy ensures that patches are sampled from the clinically relevant fat beneath the vein. Figure~\ref{fig:patch_extraction} shows both extraction regions, with pancreas and fat patches clearly delineated.

\subsection{Feature Engineering}
\label{sec:features}

For each extracted patch, we compute a feature vector combining intensity statistics and texture descriptors:

\begin{itemize}
    \item \textbf{Intensity statistics:} mean, standard deviation, and median of grayscale pixel values.
    \item \textbf{Intensity histogram:} normalized histogram with $B$ bins over the range $[0, 255]$.
    \item \textbf{Laplacian variance:} variance of the Laplacian response, capturing texture sharpness.
    \item \textbf{Local contrast:} mean of the local standard deviation computed over $3 \times 3$ neighborhoods.
    \item \textbf{Gradient magnitude:} mean of the Sobel gradient magnitude, encoding edge information.
\end{itemize}

\noindent This yields a $(6 + B)$-dimensional feature vector per patch.

\subsection{Patient-Level Classification}
\label{sec:classification}

\begin{figure*}[!t]
  \centering
  \includegraphics[width=1\textwidth]{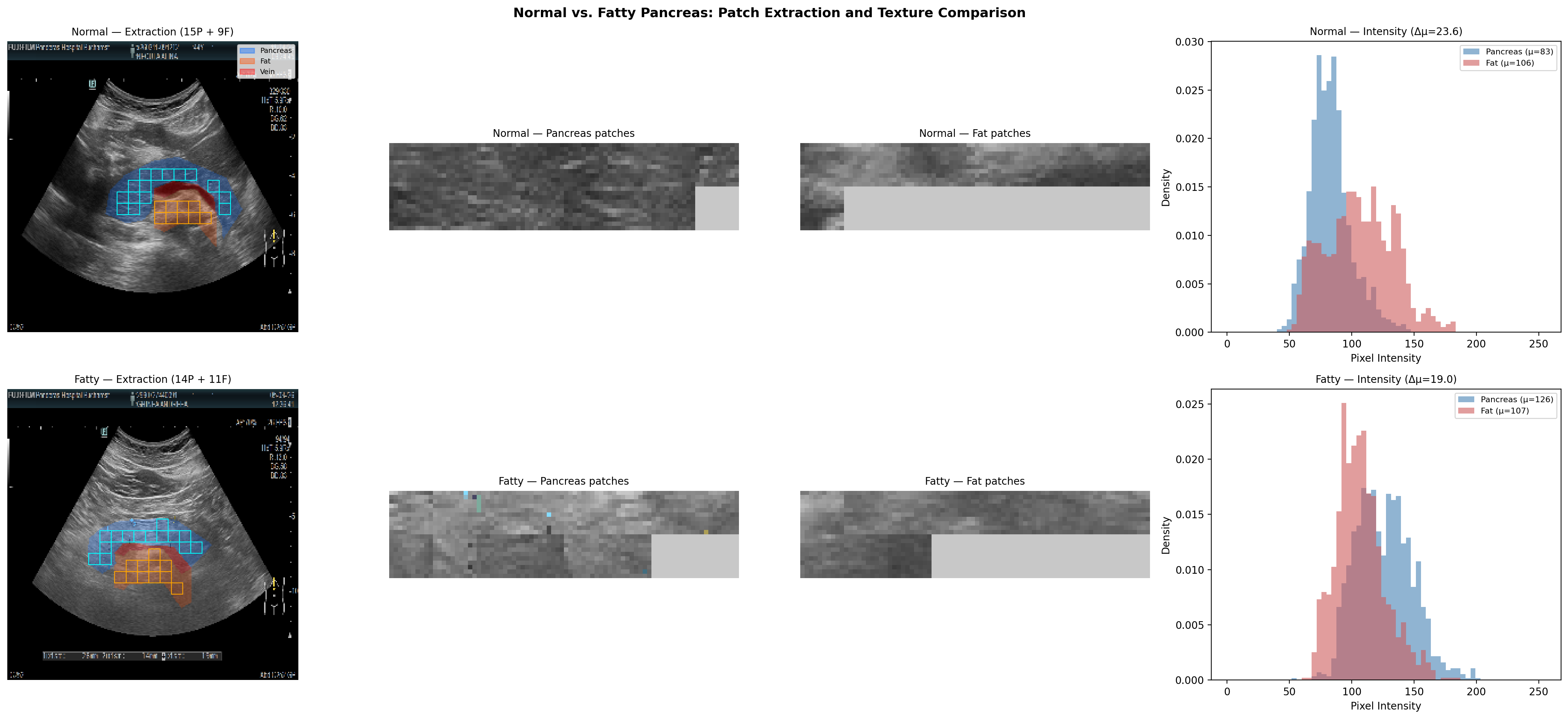}
  \caption{{\bf Comparison of patch extraction and texture profiles}.  Comparison is done between a normal pancreas (top row) and a fatty pancreas (bottom row). From left to right: extraction regions with patch grid overlay; extracted pancreas patches; extracted fat patches; pixel intensity distributions. In the normal pancreas, we can observe that the pancreas and the fat histograms are clearly separated ($\Delta\mu = 22.7$), while in the fatty pancreas, they overlap ($\Delta\mu = 19.0$), reflecting similar echogenicity, the key clinical indicator of fatty infiltration.}
  \label{fig:normal_vs_fatty}
\end{figure*}

Rather than classifying individual patches, we derive a patient-level feature vector that captures the \emph{similarity} between fat and pancreas tissue, which is the clinically relevant information. The clinical intuition is illustrated in Figure~\ref{fig:normal_vs_fatty}: in a normal pancreas, the intensity distributions of pancreas and fat patches are well-separated, whereas in a fatty pancreas the distributions overlap, reflecting similar echogenicity.

\textbf{Pairwise Distance Computation.} All patch features are standardized (zero mean, unit variance). We then compute the $L_2$ distance between every fat patch and every pancreas patch, yielding a distance matrix $D \in \mathbb{R}^{n_f \times n_p}$, where $n_f$ and $n_p$ are the number of fat and pancreas patches, respectively.

\textbf{Feature Aggregation.} The patient-level feature vector is constructed from:
\begin{itemize}
    \item Distance distribution statistics: mean, standard deviation, median, 10th and 90th percentiles of all entries in $D$ (5 features).
    \item Nearest-neighbor statistics: mean and standard deviation of the minimum distance per fat patch, and the fraction of close patch pairs below the 25th percentile (3 features).
    \item Mean feature difference between fat and pancreas patches ($6 + B$ features).
\end{itemize}

\noindent With $B{=}32$ histogram bins, this yields a 46-dimensional patient-level feature vector ($5 + 3 + 6 + 32 = 46$).

\textbf{Classification Methods.} We evaluate two paradigms:
\begin{itemize}
    \item \textbf{Unsupervised:} K-Means clustering \cite{macqueen1967kmeans} ($k{=}2$) on standardized patient features. The fatty cluster is identified as the one with lower mean pairwise distance (where fat and pancreas textures are more similar, indicating fat infiltration).
    \item \textbf{Supervised:} We train SVM \cite{cortes1995svm} (RBF and linear kernels), Random Forest \cite{breiman2001rf}, Gradient Boosting \cite{friedman2001gbm}, Logistic Regression \cite{cox1958logistic}, and K-Nearest Neighbors \cite{cover1967knn} classifiers, evaluated via 5-fold stratified cross-validation.
\end{itemize}

\section{\uppercase{Experimental Evaluation}}
\label{sec:experiments}

\subsection{Dataset}
\label{sec:dataset}

\noindent Our dataset consists of 214 abdominal ultrasound images acquired at a clinical hospital using a standard abdominal ultrasound probe. The images capture the pancreatic body region with the splenic vein visible as an anatomical landmark.

\textbf{Annotation.} All images were annotated by experts using the Labelbox platform. For each image, polygon annotations were provided for: (1) the pancreas parenchyma, and (2) the splenic vein. Additionally, 108 images were labeled at the patient level as either \textit{fatty} (53 images) or \textit{normal} (55 images) by the same experts.

\textbf{Pre-processing.} The original BMP images (1024$\times$768 pixels) were converted to PNG format and resized to 256$\times$256 for segmentation model input. Polygon annotations were transformed into binary masks for both the pancreas and splenic vein.

\textbf{Data Split and Cross-Validation Protocol.} To address the limited dataset size and provide statistically reliable performance estimates, we adopt a 5-fold cross-validation scheme for the complete end-to-end pipeline. In each fold, we draw an independent stratified 90\%/10\% random split of all 214 images (approximately 192 training, 22 validation), where stratification is applied to the labeled subset to maintain balanced fatty/normal proportions in each validation set. The two segmentation models (pancreas and splenic vein) are independently retrained from the liver-pretrained checkpoint for each fold. Segmentation metrics are computed on the fold's validation set; predicted masks are then generated for all 107 labeled patients and used to run the classification pipeline. Repeating this across five folds with different random splits yields five independent end-to-end evaluations, from which we report mean and standard deviation.

\subsection{Segmentation Results}
\label{sec:seg_results}

\textbf{Implementation Details.} Both models are trained at 256$\times$256 resolution with a batch size of 8. The liver pre-trained model is fine-tuned using Adam optimizer with learning rate $10^{-5}$. Early stopping with patience of 50 epochs is applied based on the validation Dice score. Across the five folds, the pancreas model triggers early stopping after 83--208 epochs (mean 127) and the splenic vein model after 96--153 epochs (mean 127), confirming that transfer learning from liver segmentation provides a strong initialisation that reduces the number of training iterations required.

\begin{table}[h]
\caption{{\bf Segmentation performance} ($mean\pm std$ across 5 folds, each evaluated on ${\approx}22$ held-out images). TransUNet models are independently fine-tuned from liver segmentation pre-training per fold.}\label{tab:segmentation}
\centering
\begin{tabular}{lcccc}
  \toprule
  \textbf{Target} & \textbf{Dice} & \textbf{IoU} & \textbf{Recall} & \textbf{Precision} \\
  \midrule
  P.   & \makecell{0.712 \\ {\scriptsize $\pm$0.018}} & \makecell{0.572 \\ {\scriptsize $\pm$0.018}} & \makecell{0.749 \\ {\scriptsize $\pm$0.012}} & \makecell{0.726 \\ {\scriptsize $\pm$0.031}} \\
  S.V. & \makecell{0.699 \\ {\scriptsize $\pm$0.071}} & \makecell{0.566 \\ {\scriptsize $\pm$0.072}} & \makecell{0.758 \\ {\scriptsize $\pm$0.051}} & \makecell{0.714 \\ {\scriptsize $\pm$0.092}} \\
  \bottomrule
\end{tabular}
\end{table}

Both models achieve balanced precision and recall across folds. The pancreas model shows stable cross-fold performance (Dice std of $\pm$0.018), while the splenic vein model exhibits higher variance (Dice std of $\pm$0.071), reflecting both the smaller size of the structure and the sensitivity of segmentation estimates to the composition of the small per-fold validation sets. The transfer learning approach from liver segmentation proves effective, providing a meaningful initialisation that accelerates convergence.

\begin{figure*}[!h]
  \centering
  \begin{subfigure}[b]{1\columnwidth}
    \includegraphics[width=\linewidth]{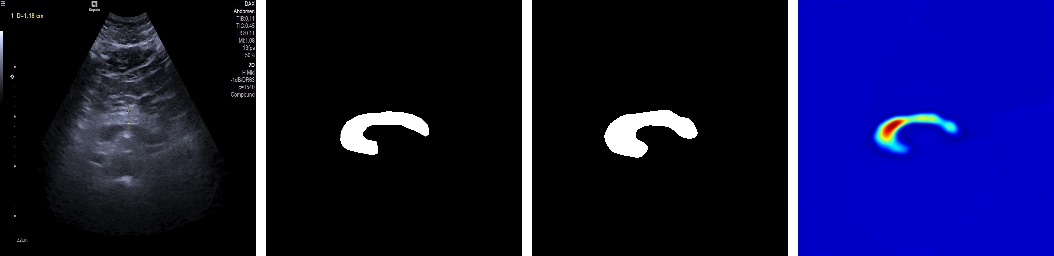}
  \end{subfigure}
  \hfill
  \begin{subfigure}[b]{1\columnwidth}
    \includegraphics[width=\linewidth]{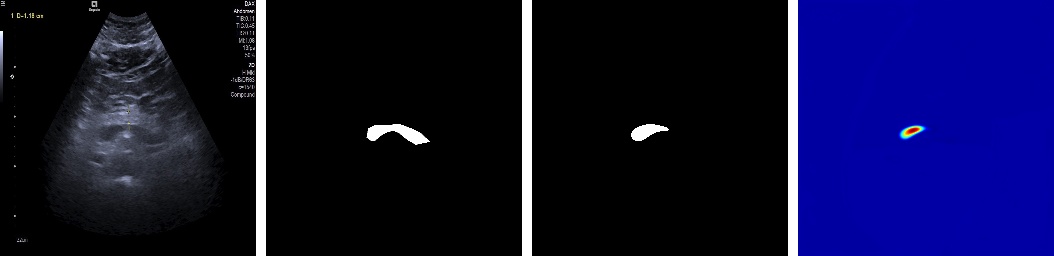}
  \end{subfigure}
  \begin{subfigure}[b]{1\columnwidth}
    \includegraphics[width=\linewidth]{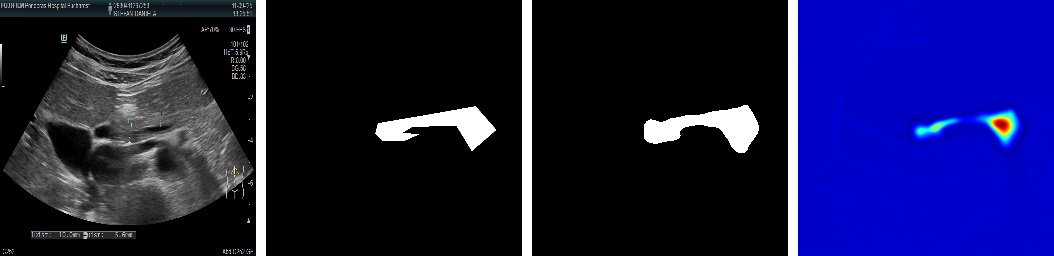}
  \end{subfigure}
  \hfill
  \begin{subfigure}[b]{1\columnwidth}
    \includegraphics[width=\linewidth]{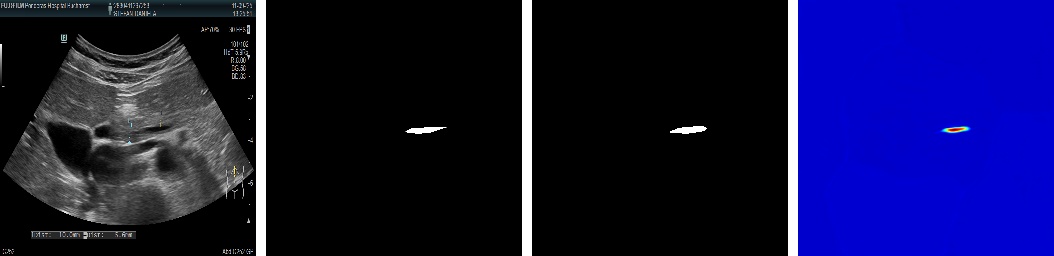}
  \end{subfigure}
    \begin{subfigure}[b]{1\columnwidth}
    \includegraphics[width=\linewidth]{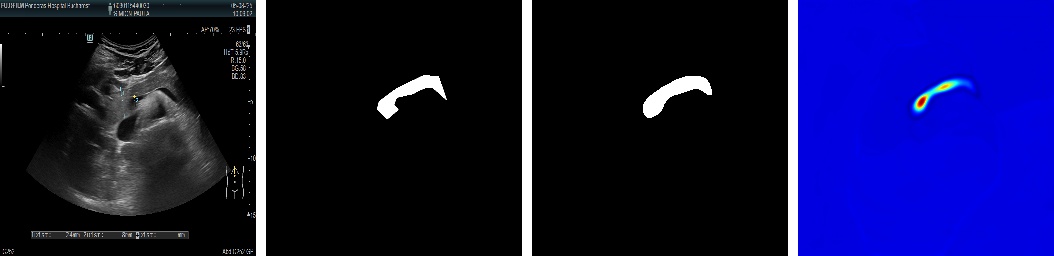}
    \caption{{\it Pancreas segmentation.}}
  \end{subfigure}
  \hfill
  \begin{subfigure}[b]{1\columnwidth}
    \includegraphics[width=\linewidth]{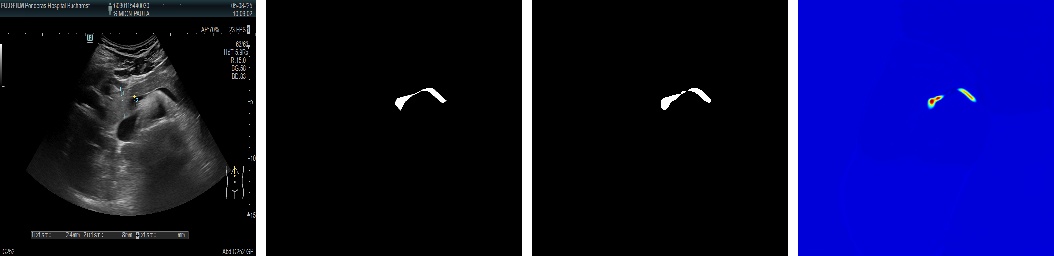}
    \caption{{\it Splenic vein segmentation.}}
  \end{subfigure}
  
  \vspace{0.3cm}
  \caption{{\bf Qualitative segmentation results}. (a) Pancreas and (b) Splenic vein segmentation of the same patient. Each panel shows, from left to right, input ultrasound image, ground-truth mask, predicted segmentation, and decoder activation heatmap.}
  \label{fig:seg_examples}
\end{figure*}

\subsubsection{Comparison with MedSAM}

We compare our TransUNet approach against MedSAM \cite{ma2024medsam}, a foundation model that adapts the Segment Anything Model (SAM) to medical imaging. We evaluate MedSAM in two settings: (1)~zero-shot, using bounding-box and foreground point prompts without any training on our data; and (2)~fine-tuned, where we fine-tune MedSAM's image encoder and mask decoder on the same training split used for TransUNet. For this direct comparison, all methods are evaluated on a common fixed 90/10 split (193 train, 21 validation), which allows a fair side-by-side comparison on identical data; the cross-validated segmentation results are reported separately in Table~\ref{tab:segmentation}. Table~\ref{tab:medsam} reports results on this fixed validation set.

\begin{table}[h]
\caption{{\bf Segmentation comparison}. TransUNet vs.\ MedSAM on a fixed 21-image validation set (common split for all methods). Zero-shot MedSAM results are on 11 validation images.}\label{tab:medsam}
\centering
\begin{tabular}{llcc}
  \toprule
  \textbf{Method} & \textbf{Target} & \textbf{Dice} & \textbf{IoU} \\
  \midrule
  \multirow{2}{*}{TransUNet (ours)} & P. & \textbf{0.780} & \textbf{0.648} \\
   & S.V. & \textbf{0.783} & \textbf{0.660} \\
  \midrule
  \multirow{2}{*}{MedSAM fine-tuned} & P. & 0.834 & 0.718 \\
   & S.V. & 0.837 & 0.725 \\
  \midrule
  \multirow{2}{*}{MedSAM zs (bbox)} & P. & 0.564 & 0.409 \\
   & S.V. & 0.530 & 0.418 \\
  \midrule
  \multirow{2}{*}{MedSAM zs (5-shot)} & P. & 0.259 & 0.156 \\
   & S.V. & 0.328 & 0.204 \\
  \midrule
  \multirow{2}{*}{MedSAM zs (50-shot)} & P. & 0.266 & 0.158 \\
   & S.V. & 0.247 & 0.154 \\
  \bottomrule
\end{tabular}
\end{table}

Zero-shot MedSAM performs poorly across all prompting strategies, with the best configuration (bounding-box prompts) achieving only 0.564 Dice for pancreas and 0.530 for splenic vein. This highlights the difficulty of pancreas segmentation in ultrasound, where low contrast, speckle noise, and anatomical variability pose challenges even for foundation models trained on diverse medical imaging data.

Fine-tuning MedSAM on our training data substantially improves performance, reaching 0.834 Dice for pancreas and 0.837 for splenic vein. Notably, these fine-tuned MedSAM results are comparable to our TransUNet approach (0.780 and 0.783 Dice, respectively). \textbf{However, this comparison has important caveats}: MedSAM requires bounding-box prompts derived from ground-truth masks at inference time, whereas our TransUNet models operate without any prompts. Furthermore, MedSAM processes images at $1024 \times 1024$ resolution compared to $256 \times 256$ for TransUNet, resulting in significantly higher computational cost. Our TransUNet approach thus offers a more practical solution for fully automated deployment where ground-truth bounding boxes are unavailable.

\subsection{Classification Results}
\label{sec:class_results}

\noindent All classification experiments in this section use \textbf{predicted segmentation masks} from the TransUNet models, demonstrating true end-to-end performance. We additionally report ground-truth mask results for comparison.

\subsubsection{Grid Search over Extraction Parameters}

We conduct a systematic grid search over 90 configurations of the key hyperparameters: patch size $s \in \{3, 5, 7, 10, 15\}$, fat region depth $\delta \in \{10, 15, 20, 30, 40, 50\}$, and histogram bins $B \in \{8, 16, 32\}$. Among configurations that evaluate all 107 labeled patients, the best unsupervised (K-Means) result is achieved with $s{=}3$, $\delta{=}15$, $B{=}32$ at 87.9\% accuracy, closely followed by $s{=}3$, $\delta{=}20$, $B{=}32$ at 87.9\% accuracy and 0.881~F1.

Smaller patch sizes ($s{=}3$) consistently outperform larger ones, as they capture finer texture details relevant to echogenicity differences. A moderate fat region depth ($\delta{=}15$--$20$) provides the best balance between sampling sufficient fat tissue and avoiding irrelevant structures. Higher histogram resolution ($B{=}32$) improves discriminability. We select $s{=}3$, $\delta{=}20$, $B{=}32$ as the operating point for subsequent experiments, as it yields the best results for both unsupervised and supervised classifiers.

\subsubsection{Classifier Comparison}

Using the selected extraction parameters ($s{=}3$, $\delta{=}20$, $B{=}32$), we compare unsupervised K-Means with six supervised classifiers. Each of the five cross-validation folds provides one full pipeline evaluation: segmentation models retrained on the fold's training split, predicted masks generated for all 107 labeled patients, and classifiers evaluated via internal 5-fold stratified cross-validation. Table~\ref{tab:classification} reports mean and standard deviation across the five folds.

\begin{table}[h]
\footnotesize
\caption{{\bf Classification performance} using predicted segmentation masks ($s{=}3$, $\delta{=}20$, $B{=}32$, $n{=}107$). Results are $mean\pm std$ across 5 independent pipeline runs with segmentation retrained per fold. K-Means is unsupervised; all other methods use internal 5-fold stratified CV. Best mean results in bold.}\label{tab:classification}
\centering
\begin{tabularx}{\linewidth}{lccccc}
  \toprule
  \textbf{Method} & \textbf{Acc} & \textbf{Prec} & \textbf{Rec} & \textbf{F1} & \textbf{$\kappa$} \\
  \midrule
  K-Means             & \makecell{0.878 \\ {\scriptsize $\pm$0.014}} & \makecell{0.849 \\ {\scriptsize $\pm$0.022}} & \makecell{0.916 \\ {\scriptsize $\pm$0.031}} & \makecell{0.881 \\ {\scriptsize $\pm$0.013}} & \makecell{0.757 \\ {\scriptsize $\pm$0.027}} \\
  \midrule
  SVM (RBF)           & \makecell{\textbf{0.897} \\ {\scriptsize $\pm$0.018}} & \makecell{0.881 \\ {\scriptsize $\pm$0.014}} & \makecell{\textbf{0.925} \\ {\scriptsize $\pm$0.028}} & \makecell{\textbf{0.898} \\ {\scriptsize $\pm$0.019}} & \makecell{\textbf{0.794} \\ {\scriptsize $\pm$0.036}} \\
  SVM (Lin)           & \makecell{0.863 \\ {\scriptsize $\pm$0.029}} & \makecell{0.870 \\ {\scriptsize $\pm$0.035}} & \makecell{0.867 \\ {\scriptsize $\pm$0.032}} & \makecell{0.862 \\ {\scriptsize $\pm$0.030}} & \makecell{0.726 \\ {\scriptsize $\pm$0.056}} \\
  Rand Forest         & \makecell{0.894 \\ {\scriptsize $\pm$0.019}} & \makecell{0.888 \\ {\scriptsize $\pm$0.017}} & \makecell{0.905 \\ {\scriptsize $\pm$0.029}} & \makecell{0.893 \\ {\scriptsize $\pm$0.021}} & \makecell{0.787 \\ {\scriptsize $\pm$0.039}} \\
  Grad Boost          & \makecell{0.858 \\ {\scriptsize $\pm$0.034}} & \makecell{0.863 \\ {\scriptsize $\pm$0.028}} & \makecell{0.853 \\ {\scriptsize $\pm$0.044}} & \makecell{0.853 \\ {\scriptsize $\pm$0.037}} & \makecell{0.715 \\ {\scriptsize $\pm$0.068}} \\
  Logistic Reg        & \makecell{0.896 \\ {\scriptsize $\pm$0.023}} & \makecell{\textbf{0.904} \\ {\scriptsize $\pm$0.021}} & \makecell{0.895 \\ {\scriptsize $\pm$0.035}} & \makecell{0.893 \\ {\scriptsize $\pm$0.025}} & \makecell{0.790 \\ {\scriptsize $\pm$0.047}} \\
  KNN ($k{=}5$)       & \makecell{0.880 \\ {\scriptsize $\pm$0.022}} & \makecell{0.858 \\ {\scriptsize $\pm$0.019}} & \makecell{0.917 \\ {\scriptsize $\pm$0.029}} & \makecell{0.883 \\ {\scriptsize $\pm$0.022}} & \makecell{0.760 \\ {\scriptsize $\pm$0.044}} \\
  \bottomrule
\end{tabularx}
\end{table}

SVM (RBF) achieves the highest mean cross-validated accuracy (89.7\%\,$\pm$\,1.8\%) and F1 (0.898\,$\pm$\,0.019), with Cohen's $\kappa{=}0.794\,\pm\,0.036$ indicating substantial agreement with expert labels. Logistic Regression and Random Forest perform comparably (89.6\% and 89.4\% mean accuracy respectively) with similarly low standard deviations, confirming stable performance across folds. The unsupervised K-Means baseline achieves 87.8\%\,$\pm$\,1.4\% accuracy, demonstrating that the feature engineering captures meaningful clinical signal even without labeled training data.

Gradient Boosting exhibits the highest variance ($\pm$3.4\% accuracy, $\kappa$ std of $\pm$0.068), suggesting sensitivity to the specific training split at this dataset scale. In contrast, SVM (RBF) and Random Forest show the most stable behaviour. The low standard deviations across all methods (below $\pm$3.5\% accuracy) confirm that results are robust and not an artefact of a particular data split.

\subsubsection{Predicted vs.\ Ground-Truth Masks}

A key question is whether the classification pipeline is robust to segmentation errors. Table~\ref{tab:gt_vs_pred} compares classification performance when using ground-truth (GT) versus predicted segmentation masks with identical extraction parameters.

\begin{table}[h]
\caption{{\bf Classification with ground-truth vs.\ predicted masks} ($s{=}3$, $\delta{=}20$, $B{=}32$, $n{=}107$).}\label{tab:gt_vs_pred}
\centering
\begin{tabular}{lcccc}
  \toprule
  \textbf{Method} & \multicolumn{2}{c}{\textbf{GT}} & \multicolumn{2}{c}{\textbf{Predictions}} \\
  \cmidrule(lr){2-3} \cmidrule(lr){4-5}
   & Acc & F1 & Acc & F1 \\
  \midrule
  K-Means             & 0.860 & 0.870 & 0.879 & 0.881 \\
  \midrule
  SVM (RBF)           & 0.907 & 0.909 & 0.925 & 0.925 \\
  Rand Forest         & 0.916 & 0.914 & 0.907 & 0.906 \\
  Logistic Regr       & 0.926 & 0.921 & 0.925 & 0.925 \\
  KNN ($k{=}5$)       & 0.907 & 0.909 & 0.906 & 0.910 \\
  \bottomrule
\end{tabular}
\end{table}

The results demonstrate notable robustness: classification performance with predicted masks is comparable to, and in some cases slightly exceeds, that with ground-truth masks. Logistic Regression achieves 92.5\% accuracy with predicted masks versus 92.6\% with ground-truth masks, a negligible difference. K-Means and SVM~(RBF) actually improve slightly with predicted masks (87.9\% vs.\ 86.0\% and 92.5\% vs.\ 90.7\%, respectively).

Several factors may explain this counterintuitive result. First, the predicted masks may introduce a boundary-smoothing effect that reduces noise from annotation inconsistencies: manual polygon annotations in ultrasound are imprecise due to ambiguous tissue boundaries. Second, the predicted masks exhibit consistent error patterns across patients (systematic bias), whereas ground-truth annotations may contain characteristic variations despite originating from the same expert team. Third, and most importantly, the pairwise texture comparison framework measures \emph{relative} differences between fat and pancreas regions rather than absolute texture values, making it tolerant to moderate shifts in region boundaries.

This result also implies that further improvements in segmentation accuracy alone may not directly translate to better classification, and that the classification bottleneck likely lies in dataset size or feature representation rather than segmentation precision.


\subsubsection{Feature Space Visualization}

\begin{figure}[!h]
  \centering
  \includegraphics[width=7.7cm]{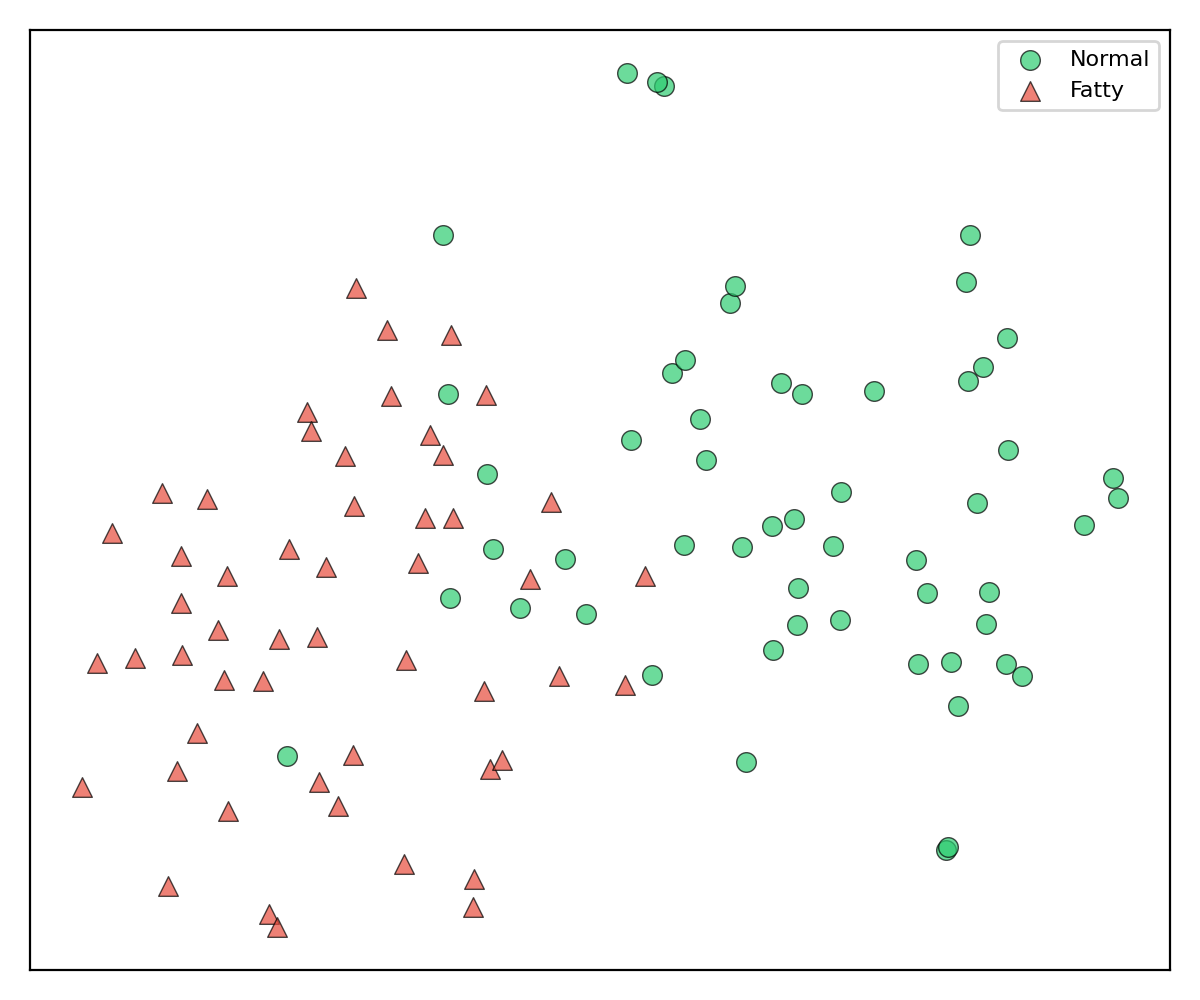}
  \caption{{\bf t-SNE visualization of patient features colored by GT labels}. Fatty pancreases (triangles, red) and normal pancreases (circles, green) form largely distinct clusters.}
  \label{fig:tsne}
\end{figure}

\begin{figure}[!h]
  \centering
  \includegraphics[width=7.7cm]{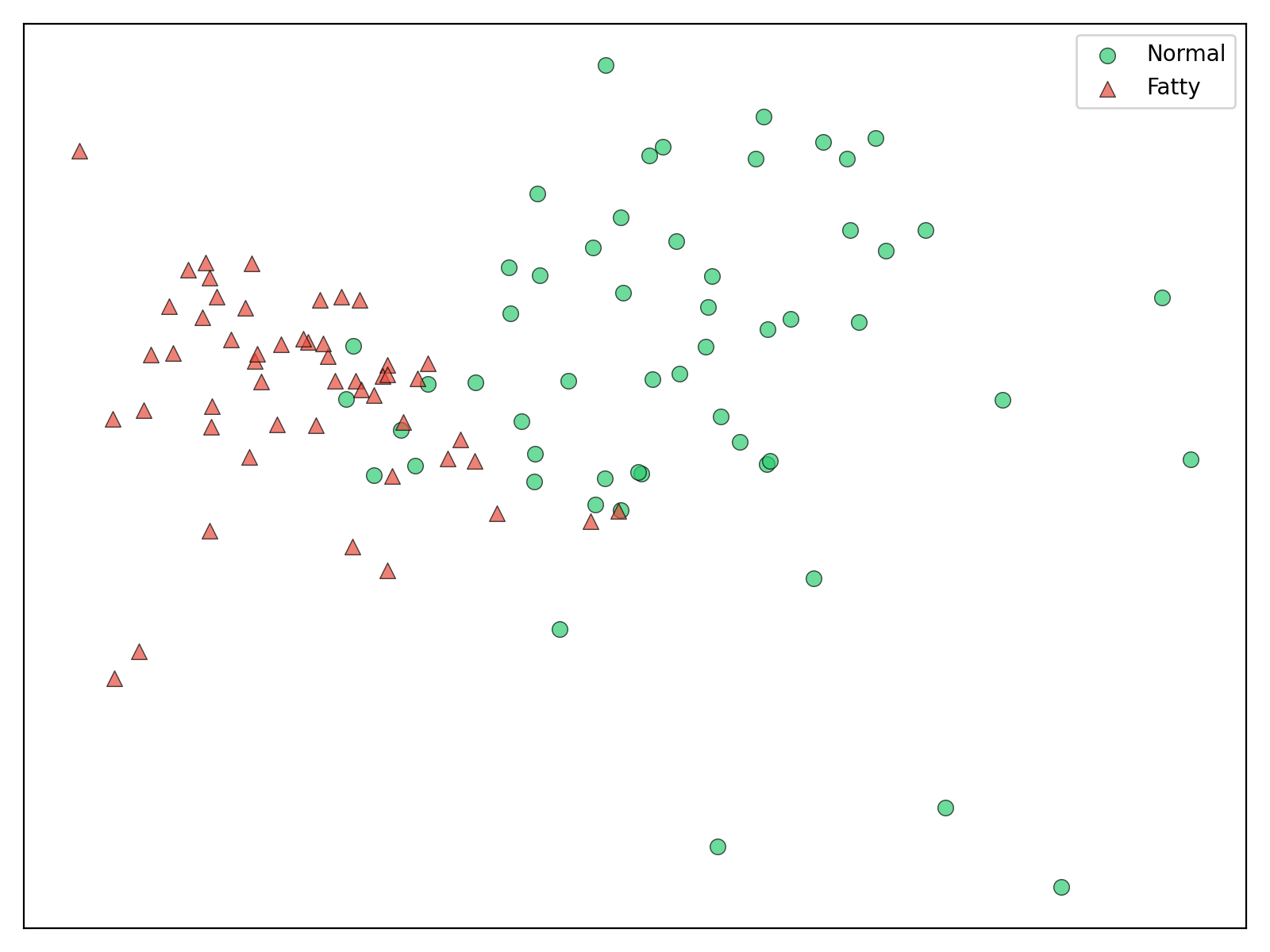}
  \caption{{\bf PCA projection of patient features}. The two classes show clear separation along the first principal component (x-axis displays $28.7\%$ variance explained).}
  \label{fig:pca}
\end{figure}

Figure~\ref{fig:tsne} shows a t-SNE projection of the 46-dimensional patient feature vectors, colored by ground-truth labels. The visualization reveals a clear separation between fatty and normal patients, with a small overlap zone corresponding to borderline cases. The PCA projection (Figure~\ref{fig:pca}) shows that the first two principal components capture 28.7\% and 14.7\% of the variance, with the fatty and normal clusters well-separated along the first component.

\subsection{Parameter Sensitivity Analysis}
\label{sec:sensitivity}

\begin{figure}[!h]
  \centering
  \includegraphics[width=8.2cm]{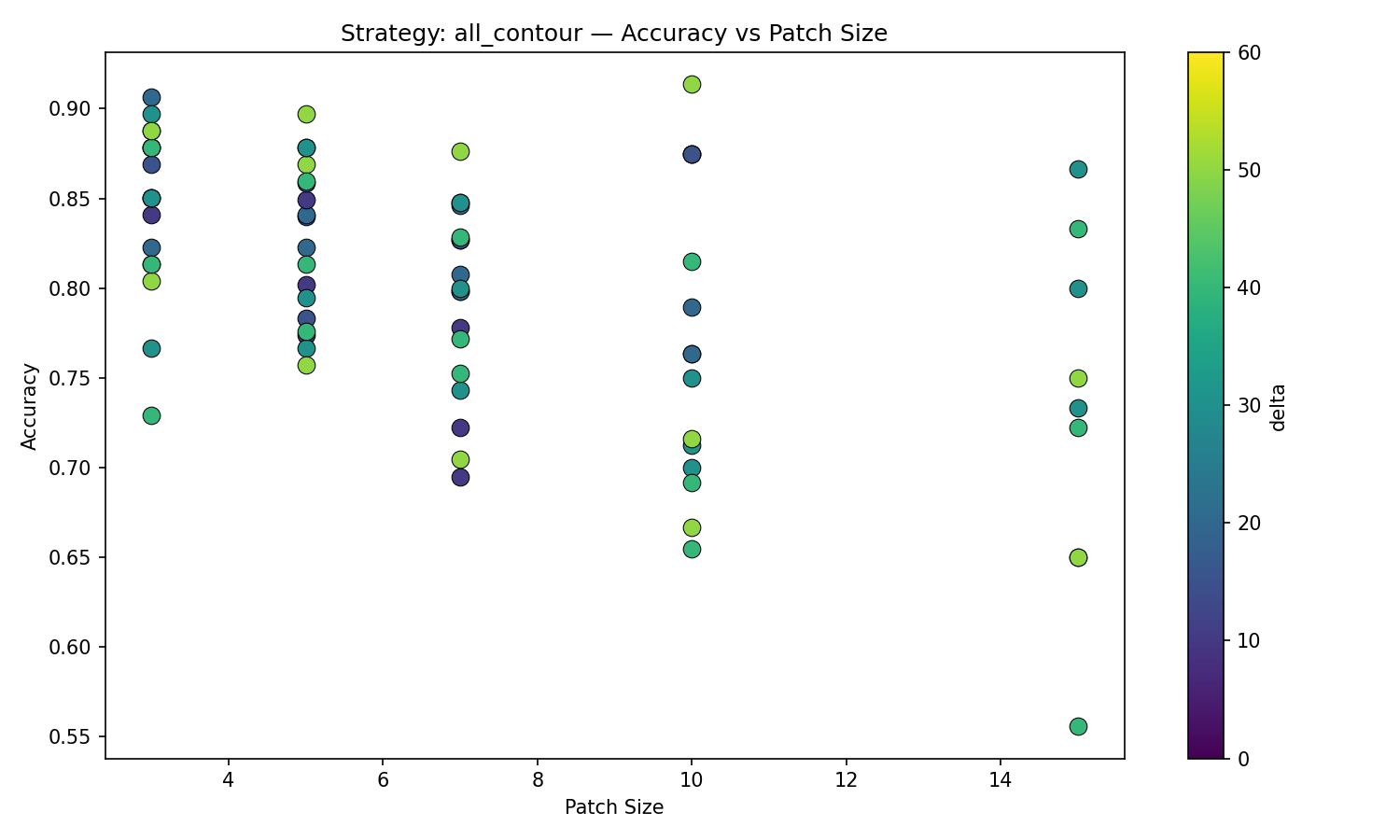}
  \caption{{\bf Sensitivity analysis}. K-Means classification accuracy as a function of patch size and fat region depth $\delta$, with $B{=}32$ histogram bins.}
  \label{fig:sensitivity}
\end{figure}

Figure~\ref{fig:sensitivity} illustrates the sensitivity of classification accuracy to key parameters. The analysis reveals:

\begin{itemize}
    \item \textbf{Patch size:} Accuracy decreases as patch size increases beyond $3{\times}3$, likely because larger patches average out the fine-grained texture differences. Very large patches ($15{\times}15$) also reduce the number of extractable patches, limiting the analysis to fewer patients.
    \item \textbf{Fat region depth ($\delta$):} Moderate values ($\delta{=}15$--$20$) yield the best results. Too small values provide insufficient fat tissue, while large values may sample tissue far from the vein that is less diagnostically relevant.
    \item \textbf{Histogram bins ($B$):} Higher resolution ($B{=}32$) consistently outperforms lower resolutions, providing finer texture discrimination.
\end{itemize}

\subsection{Experimental Setup}
All our experiments were conducted on a Google Colab NVIDIA T4 GPU (16GB VRAM, Turing architecture with Tensor Cores), while inference was performed on CPU using a MacBook Air M1 with 8GB RAM, reflecting a relatively modest hardware setup.

\textbf{Runtime.} The full pipeline processes 214 images in 112\,s on CPU (0.52\,s/image): 1.2\,s for model loading, 110.8\,s for segmentation and feature extraction, and 0.3\,s for classification.

\section{\uppercase{Conclusions and Future Work}}
\label{sec:conclusion}

\noindent We presented an end-to-end framework for automatic classification of normal versus fatty pancreas from abdominal ultrasound images. Our pipeline integrates three key components: TransUNet-based segmentation of the pancreas and splenic vein via transfer learning (mean cross-validated Dice of $0.712\,\pm\,0.018$ for pancreas and $0.699\,\pm\,0.071$ for splenic vein), anatomically-guided patch extraction mimicking clinical assessment protocols, and patient-level classification through pairwise texture comparison.

Our experiments on 214 clinical ultrasound images with 107 expert-labeled cases demonstrate that the proposed feature engineering effectively captures the clinically relevant echogenicity differences. Under 5-fold cross-validation with independent segmentation retraining per fold, SVM (RBF) achieves a mean accuracy of 89.7\%\,$\pm$\,1.8\% and F1 of 0.898\,$\pm$\,0.019, with Cohen's $\kappa{=}0.794\,\pm\,0.036$ indicating substantial agreement with expert radiologists. The unsupervised K-Means baseline reaches 87.8\%\,$\pm$\,1.4\% accuracy. The consistently low standard deviations across all methods confirm that the results are stable and reliable despite the limited dataset size.

We additionally showed that zero-shot MedSAM performs poorly on this task (Dice 0.564 for pancreas), while fine-tuning MedSAM on our data yields competitive segmentation results (Dice 0.834). However, unlike our TransUNet approach, MedSAM requires ground-truth bounding-box prompts at inference time, making it less suitable for fully automated deployment.

\textbf{Limitations.} Several limitations should be acknowledged. The dataset, while expertly annotated, is relatively small (214 images, 107 labeled for classification) and originates from a single clinical center, which may limit generalizability. All images were annotated by the same expert team; inter-observer variability in both segmentation and classification labels was not assessed. While the 5-fold cross-validation strategy mitigates overfitting concerns and provides reliable classification estimates, the individual segmentation validation sets contain only approximately 21 images each, providing limited statistical power for segmentation performance estimation. The binary classification (normal vs.\ fatty) does not capture the full clinical spectrum of fatty infiltration severity. Finally, no external validation on an independent dataset was performed.

\textbf{Future Work.} Several directions merit further investigation:
\begin{itemize}
    \item Developing a prompt-free MedSAM variant or integrating automatic bounding-box prediction to enable fully automated MedSAM-based segmentation.
    \item Extracting deep features from the segmentation model's intermediate layers as an alternative to handcrafted texture features.
    \item Extending from binary (normal/fatty) to multi-grade fatty pancreas assessment.
    \item Expanding the dataset with multi-center data for improved generalizability and external validation.
\end{itemize}

\section{\uppercase{Implementation}}

The code is anonymously available at: \href{https://github.com/anon915/Unified-Framework-for-Detection-and-Classification-of-Fatty-Pancreas-in-Ultrasound-Images}{this link}.

\section*{\uppercase{Acknowledgements}}


\end{document}